\documentclass[11pt]{article}
\usepackage{acl}
\usepackage{times}
\usepackage{latexsym}
\usepackage[T1]{fontenc}
\usepackage[utf8]{inputenc}
\usepackage{microtype}
\newcommand{\hflogo}{%
  \raisebox{-3pt}{\includegraphics[height=2em]{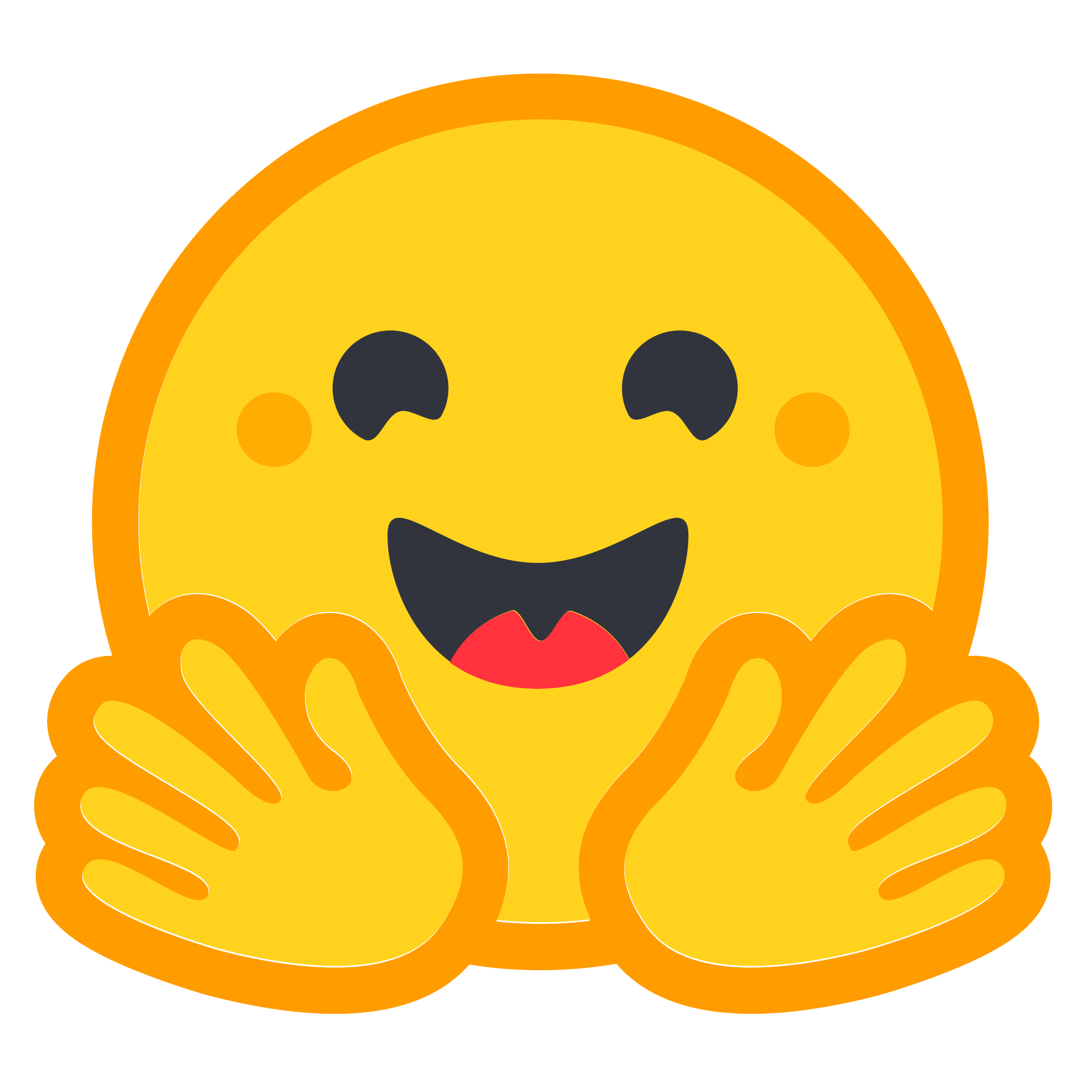}}%
}

\newcommand{\githublogo}{%
  \raisebox{-3pt}{\includegraphics[height=2em]{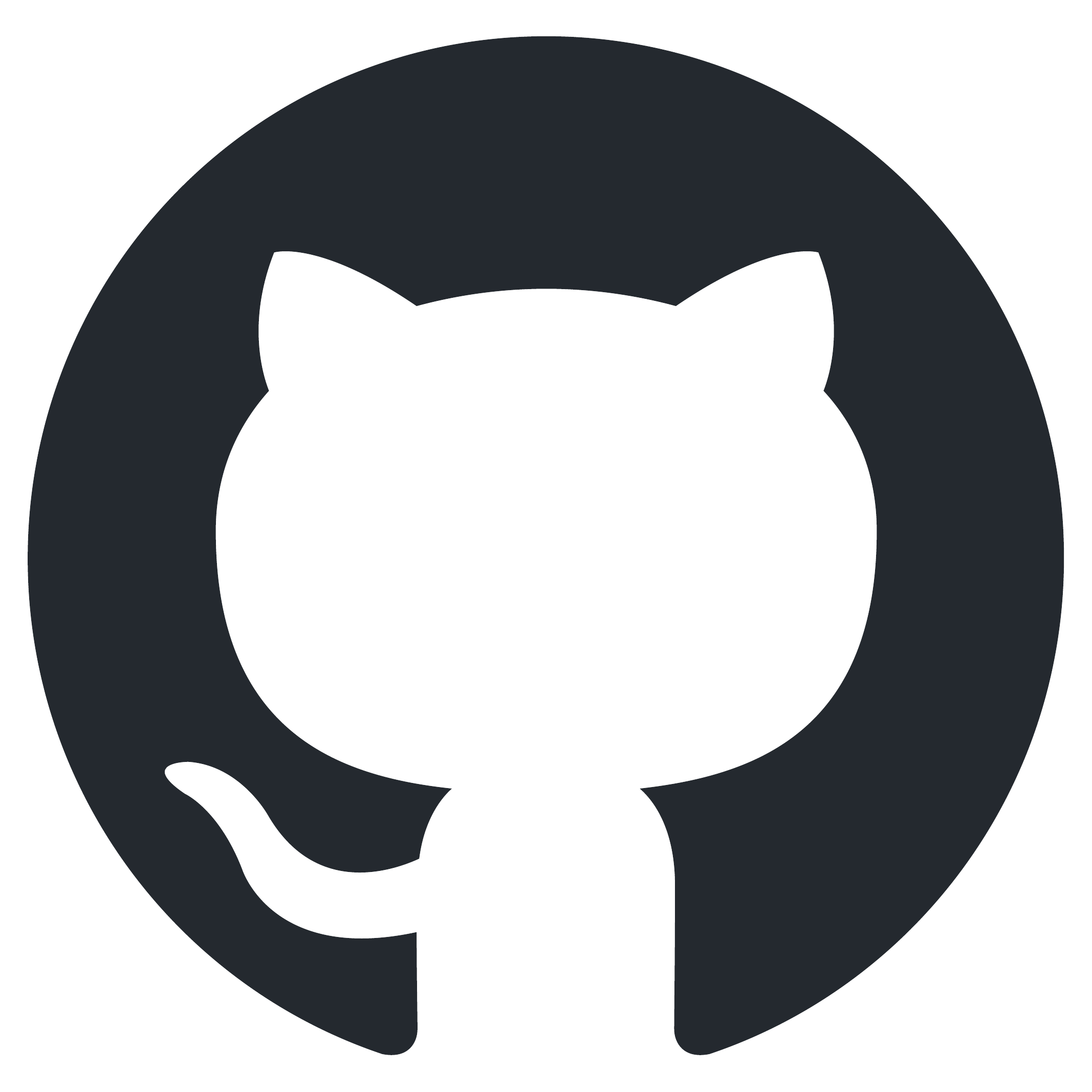}}%
}
\usepackage{graphicx}
\usepackage{booktabs}
\usepackage{multirow}
\usepackage{array}
\usepackage{pifont}
\usepackage{xcolor}
\usepackage{tikz}
\usepackage{float}
\usetikzlibrary{positioning,arrows.meta,shapes.geometric,calc,fit,backgrounds}

\definecolor{lgInk}{HTML}{000000}      
\definecolor{lgIOline}{HTML}{222222}   
\definecolor{lgIOfill}{HTML}{FFFFFF}   
\definecolor{lgArrow}{HTML}{111111}    

\definecolor{lgStage1}{HTML}{FF0055}   
\definecolor{lgStage2}{HTML}{FF9900}   
\definecolor{lgStage3}{HTML}{00CC33}   
\definecolor{lgStage4}{HTML}{0066FF}

\title{Lipi-Ghor: A Large-Scale, Diarized, Multi-Domain Bengali Speech Corpus and a Scalable Pipeline for Building It}

\usepackage{orcidlink}

\author{
  Sanjid Hasan\orcidlink{0009-0006-1271-9786}$^{1}$ \quad
  A H M Fuad\orcidlink{0009-0008-1716-8754}$^{2}$ \quad
  Risalat Labib\orcidlink{0009-0003-2399-4204}$^{2}$ \quad
  Bayazid Hasan\orcidlink{0009-0007-7108-225X}$^{2}$ \\[4pt]
  $^{1}$Department of Electrical and Electronic Engineering, KUET, Khulna, Bangladesh \\
  $^{2}$Department of Computer Science and Engineering, BUET, Dhaka, Bangladesh \\[4pt]
  {\tt \{sanjid9657, ahmfuad9, risalatlabib, bayazid741854\}@gmail.com}
}

\begin{document}

\maketitle

\begin{abstract}
\vspace{-1.0pt}
Bengali, spoken by over 230 million people, remains severely under-resourced in spoken language research.
Existing open Bengali ASR corpora are confined to short, read or crowd-sourced speech and lack speaker diarization and long-form audio.
We present \textbf{Lipi-Ghor}, a large-scale weakly-supervised Bengali speech corpus comprising 885 hours of audio sourced from 998 YouTube videos across 596 unique channels, spanning 17 diverse domains including audiobooks, drama, religious sermons, news, and regional dialects.
Each recording is annotated in SSTT format (Speaker, Speech, Transcription, Timestamp), which combines automated diarization with Bengali YouTube caption tracks.
To our knowledge, Lipi-Ghor is the first Bengali speech corpus to pair large-scale coverage with speaker-diarized, long-form audio annotations, both of which are absent from existing open Bengali resources.
We also present the construction methodology as a \emph{replicable pipeline} for any language with a sufficient captioned YouTube presence. Targeted channel curation across a handful of rich domains yields corpus scales that were previously out of reach for low-resource languages.
We benchmark four widely-used ASR models under zero-shot and fine-tuned conditions. All four degrade consistently on emotive, dialectal, and long-form content.
Because caption targets are noisy, we also release a human-annotated gold evaluation split, verified by 17 volunteer annotators over 30 hours of human-in-the-loop correction and stratified across all 17 domains.
We release our original contributions (video-ID lists, diarization output, pipeline code, and the gold annotations) under CC-BY-4.0, and distribute caption-derived transcripts for research use only; no raw audio is redistributed.
\footnote{%
\scriptsize
\href{https://huggingface.co/datasets/Sanjidh090/Lipi-Ghor-bn-882-SSTT}
{\hflogo\, \textsc{Dataset}}
\qquad
\href{https://github.com/Sanjidh090/Lipi-Ghor-bn-882-SSTT}
{\githublogo\, \textsc{Code}}
}
\end{abstract}

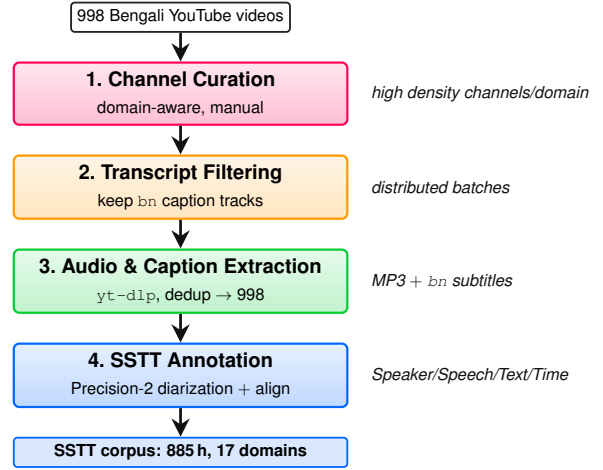
\begin{figure}[t!]
\centering
\resizebox{\columnwidth}{!}{%
\begin{tikzpicture}[
  font=\sffamily,
  node distance=5.5mm,
  stage/.style={
    rectangle, rounded corners=3pt, draw=#1, line width=1.2pt, 
    top color=#1!10, bottom color=#1!25, 
    text=lgInk, align=center, inner sep=5pt,
    minimum width=6.4cm, minimum height=11mm},
  io/.style={
    rectangle, rounded corners=2pt, draw=lgIOline, line width=1pt,
    fill=lgIOfill, text=lgInk, align=center, inner sep=3pt,
    font=\sffamily\small, minimum width=4.2cm},
  tag/.style={
    anchor=west, font=\sffamily\footnotesize\itshape, text=lgArrow, xshift=3mm},
  flow/.style={-{Stealth[length=3.5mm,width=3.5mm]}, line width=1.5pt, draw=lgArrow}, 
]

\node[io] (in) {\strut 998 Bengali YouTube videos};

\node[stage=lgStage1, below=of in] (s1)
  {\textbf{1.\ Channel Curation}\\[1pt]\footnotesize domain-aware, manual};
\node[stage=lgStage2, below=of s1] (s2)
  {\textbf{2.\ Transcript Filtering}\\[1pt]\footnotesize keep \texttt{bn} caption tracks};
\node[stage=lgStage3, below=of s2] (s3)
  {\textbf{3.\ Audio \& Caption Extraction}\\[1pt]\footnotesize \texttt{yt-dlp}, dedup $\rightarrow$ 998};
\node[stage=lgStage4, below=of s3] (s4)
  {\textbf{4.\ SSTT Annotation}\\[1pt]\footnotesize Precision-2 diarization $+$ align};

\node[io, below=of s4, fill=lgStage4!15, draw=lgStage4, text=lgInk,
      font=\sffamily\small\bfseries, minimum width=6.4cm] (out)
  {\strut SSTT corpus: 885\,h, 17 domains};

\draw[flow] (in) -- (s1);
\draw[flow] (s1) -- (s2);
\draw[flow] (s2) -- (s3);
\draw[flow] (s3) -- (s4);
\draw[flow] (s4) -- (out);

\node[tag] at (s1.east) {high density channels/domain};
\node[tag] at (s2.east) {distributed batches};
\node[tag] at (s3.east) {MP3 $+$ \texttt{bn} subtitles};
\node[tag] at (s4.east) {Speaker/Speech/Text/Time};

\end{tikzpicture}%
}
\caption{The Lipi-Ghor construction pipeline. Four stages turn a curated set of Bengali YouTube channels into a speaker-diarized, SSTT-annotated corpus. The methodology is language-agnostic; only Stage~1 curation is language-specific.}
\label{fig:pipeline}
\end{figure}

\section{Introduction}

Automatic Speech Recognition (ASR) for low-resource languages is bottlenecked by data.
English ASR now approaches near-human performance, but languages such as Bengali, the seventh most spoken in the world with over 230 million native speakers, remain severely underserved.
Existing open Bengali ASR resources are narrow in domain and confined to short read or crowd-sourced utterances. They lack diarization, speaker annotations, and long-form audio, which limits their use for training modern speech-language models.

Real Bengali speech makes the problem harder.
Unlike English or Mandarin, Bengali carries substantial dialectal variation, code-switching with English, register shifts across formal news broadcast, informal Natok drama, and Islamic religious sermons, and high speaker density in long-form audio.
Corpora built from read speech or studio recordings miss this diversity, and the models trained on them degrade sharply in the wild.

We address this gap with \textbf{Lipi-Ghor} (``House of Scripts''), a large-scale Bengali speech corpus built from YouTube.
To our knowledge, it is the first Bengali corpus to combine large-scale coverage with speaker-diarized, long-form audio in a single resource.
We make three contributions.
First, we describe a scalable, replicable data construction pipeline that uses domain-aware channel curation, caption-track filtering, distributed transcript fetching, and automated diarization to produce SSTT-annotated corpora at scale.
Second, we instantiate this pipeline for Bengali, yielding 885 hours of speech across 17 domains from 596 unique channels. A small number of well-chosen channels per domain yields hundreds of hours of training-ready data, which bears directly on corpus construction for other under-resourced languages.
Third, we release a human-annotated gold evaluation split and benchmark four widely-used ASR systems in zero-shot and fine-tuned settings, with a per-domain and duration-based analysis of where they degrade.

Figure~\ref{fig:pipeline} provides an overview of the Lipi-Ghor construction process.

\section{Related Work}

\paragraph{Bengali ASR Datasets.}
Prior open Bengali ASR resources are limited in scale, register, and annotation richness, and none pair large-scale training audio with speaker diarization and long-form recordings.
Early corpora such as SHRUTI \citep{das2011shruti} contain only tens of hours of read speech from a handful of speakers.
Common Voice Bengali \citep{alam2022bengalicv,ardila-etal-2020-common} provides several hundred hours of crowd-sourced read speech but covers only a narrow, sentence-level acoustic register.
The BengaliAI Speech Corpus \citep{bengaliai2023} offers broader, larger-scale coverage but was collected under controlled crowd-sourced conditions without speaker-attributed long-form audio.
Large Indic multilingual efforts include the Bengali portions of IndicSUPERB/Kathbath \citep{javed2023indicsuperb}, IndicWav2Vec \citep{javed2022indicwav2vec}, and the Vistaar benchmark suite \citep{bhogale2023vistaar}, but these remain sentence-level and lack conversational diarization.
OOD-Speech \citep{oodsp2023} introduced the first out-of-distribution benchmark for Bengali, with 23 hours of manually annotated test data across 17 sources, but it is a test set, not a diarized training corpus.
BengaliLoop \citep{bengaliloop2025} released a long-form diarization benchmark with human-in-the-loop transcripts, but at evaluation scale.
Table~\ref{tab:corpora} situates Lipi-Ghor against prior Bengali resources.

\paragraph{Web- and YouTube-Based Corpus Construction.}
The People's Speech \citep{galvez2021peoples} and GigaSpeech \citep{chen2021gigaspeech} assembled tens of thousands of hours of English from web sources, while VoxPopuli \citep{wang2021voxpopuli} scaled a similar approach to 23 European languages.
Most directly related, YODAS \citep{li2023yodas} harvested over 500k hours from YouTube subtitles across more than 100 languages,only 27 hours for bangla language.
These efforts are either English-centric or provide only segment-level transcripts without speaker diarization, and the Bengali material they contain is sparse.
Two problems recur in such pipelines: filtering for target-language caption quality, and managing API rate constraints at scale. We address both in our pipeline design (\S\ref{sec:pipeline}).

\paragraph{Speaker Diarization Corpora.}
Diarization benchmarks such as AMI \citep{carletta2005ami}, DIHARD~III \citep{ryant2021dihard3}, and VoxConverse \citep{chung2020voxconverse} establish the difficulty of speaker attribution on naturalistic audio, but none cover Bengali. Lipi-Ghor brings diarized annotation to Bengali at training scale.

\paragraph{Weakly Supervised ASR.}
Weakly supervised learning from noisy web transcripts is well established \citep{whisper2022}, as are massively multilingual models such as MMS \citep{pratap2024mms} and SeamlessM4T \citep{seamless2023m4t}.
Auto-generated captions exhibit systematic error patterns distinct from human transcription \citep{kafle2017evaluating}, a limitation we address through our gold evaluation split.

\begin{table}[t]
\centering
\small
\setlength{\tabcolsep}{4pt}
\begin{tabular}{lrccc}
\toprule
\textbf{Corpus} & \textbf{Hrs} & \textbf{Dom.} & \textbf{Diar.} & \textbf{Long} \\
\midrule
Common Voice bn      & $\sim$40 & 1  & \ding{55} & \ding{55} \\
BengaliAI Speech     & $\sim$1200$^\ast$ & few & \ding{55} & \ding{55} \\
OOD-Speech (test)    & 23 & 17 & \ding{55} & \ding{55} \\
BengaliLoop (eval)   & small & few & \ding{51} & \ding{51} \\
\midrule
\textbf{Lipi-Ghor}   & \textbf{885} & \textbf{17} & \ding{51} & \ding{51} \\
\bottomrule
\end{tabular}
\caption{Open Bengali speech resources by scale, domain count, speaker diarization, and long-form audio. $^\ast$Predominantly short, read/controlled speech.}
\label{tab:corpora}
\end{table}

\section{The Lipi-Ghor Pipeline}
\label{sec:pipeline}

Our construction pipeline consists of four stages: domain-aware channel curation, transcript availability filtering, audio and caption extraction, and SSTT annotation.

\subsection{Stage 1: Domain-Aware Channel Curation}

Rather than crawling YouTube indiscriminately, we perform targeted manual curation of Bengali YouTube channels across predefined domain categories.

The initial collection process focused on identifying high-volume channels within each domain to efficiently bootstrap data acquisition. These anchor channels provided substantial amounts of domain-relevant speech, after which additional channels were incorporated to improve speaker, topic, and recording diversity. The final corpus therefore combines contributions from both high-volume sources and a broader long-tail distribution of channels.
The bottleneck is not data collection but \emph{domain-aware channel discovery}, that is, finding which channels have enough Bengali-language caption coverage.

\subsection{Stage 2: Transcript Availability Filtering}

Only videos with an existing Bengali caption track (language code \texttt{bn}) are retained.
This filtering uses our \texttt{ytranscript} toolkit,\footnote{%
\href{https://github.com/Sanjidh090/ytranscript}{\githublogo\,\texttt{Sanjidh090/ytranscript}}%
} which queries caption availability without downloading audio.
To manage YouTube API rate constraints at scale, video IDs are split into batches and processed in parallel across multiple compute environments (Kaggle notebooks, Google Colab sessions, GitHub Codespaces), with results merged after each batch.

\subsection{Stage 3: Audio and Caption Extraction}

Audio (MP3) and Bengali subtitle tracks are extracted using \texttt{yt-dlp}.
Only Bengali-track captions are retained; auto-translated tracks are discarded.
The extraction stage produced 1,019 audio-caption recordings. Cross-domain deduplication removed duplicate content shared across channels, yielding 998 unique videos that constitute the released corpus.

\subsection{Stage 4: SSTT Annotation}

Every audio file is diarized with Precision-2, a speaker-diarization model from pyannoteAI \citep{pyannoteprecision2}, built on the \texttt{pyannote.audio} framework \citep{bredin2020pyannote}.
Speaker count is inferred per recording; labels are local to each video and not linked across videos.
Precision-2 output is aligned with Bengali caption transcripts to produce SSTT (Speaker, Speech, Transcription, Timestamp) annotations at the segment level.
All 1,019 files passed integrity checks: no invalid intervals, no duplicate segments, no annotation extending past recording end.
Because Precision-2 is a commercial, API-gated model, we release its output so that the corpus can be used without re-running diarization.
Final data is packaged into three splits: (i)~raw diarization output, (ii)~diarization aligned with transcription (885-hour SSTT corpus), and (iii)~segment-chunked ASR-ready data (364 videos pre-chunked for direct training use).

\section{Dataset Description}

\subsection{Scale and Coverage}

Table~\ref{tab:summary} summarizes the released corpus after
cross-domain deduplication. Per-domain breakdowns appear in
Appendix~\ref{app:stats}.
The corpus is dominated by audiobooks (254h), talkshows (142h), and movies (122h), with smaller but linguistically important contributions from regional dialects, religious sermons, and children's cartoons.

\begin{figure}[t]
\centering
\includegraphics[width=\columnwidth]{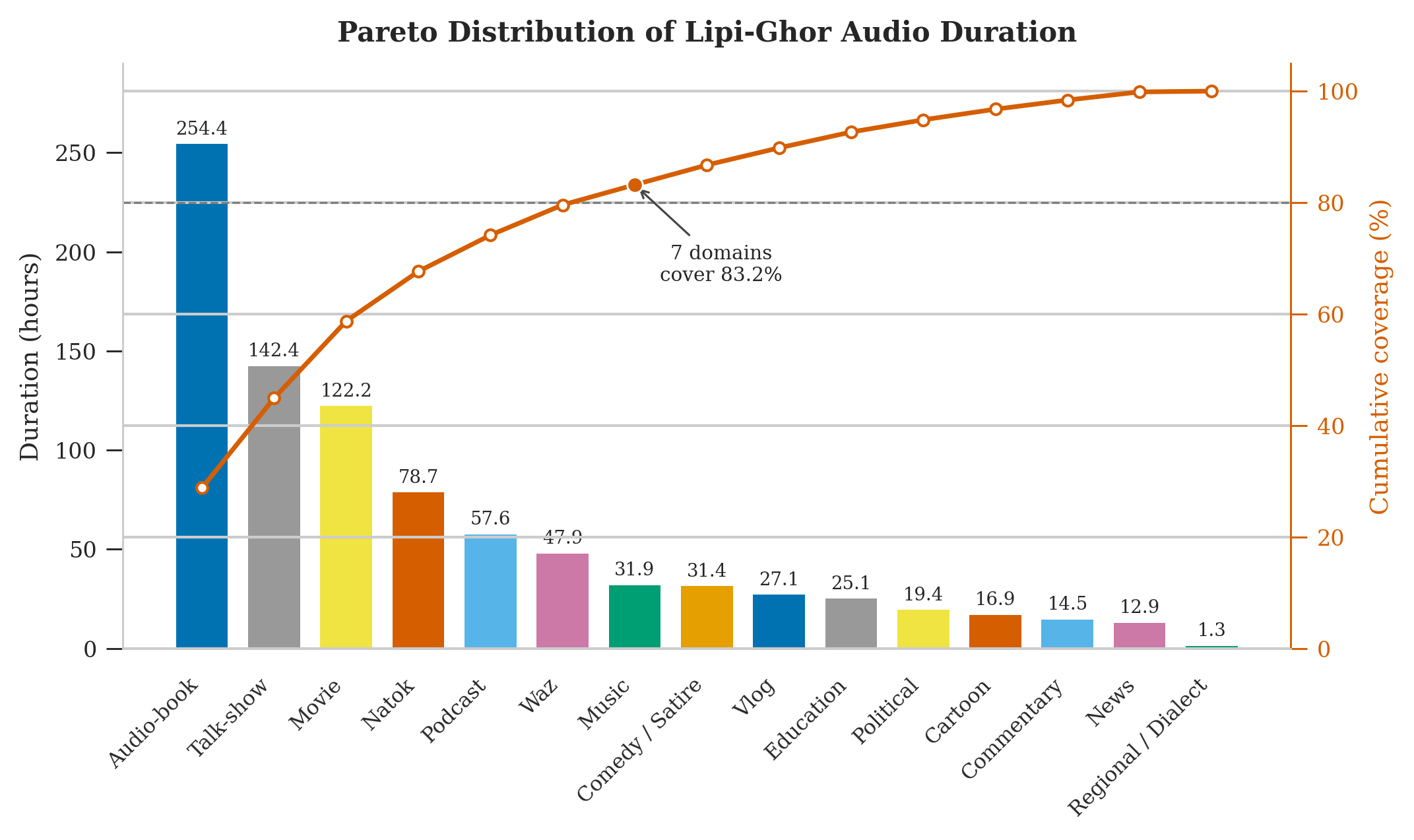}
\caption{Pareto view of audio duration by domain. Seven of the seventeen domains account for roughly 83\% of the corpus, reflecting where Bengali caption coverage is densest on YouTube.}
\label{fig:pareto}
\end{figure}

The distribution is heavily skewed, as Figure~\ref{fig:pareto} makes clear: three domains account for 58.6\% of audio, and seven of seventeen reach 83.0\%.
This shape is a direct consequence of the curation strategy: audiobook and talkshow channels upload long, caption-rich episodes, whereas dialect and health content on YouTube is short-form and rarely captioned.
The skew means domain balance is a function of what is available with Bengali captions, not of what would be ideal for training.

\begin{table}[t]
\centering
\small
\begin{tabular}{lr}
\toprule
\textbf{Property} & \textbf{Value} \\
\midrule
Total Videos        & 998 \\
Unique Channels     & 596 \\
Total Duration      & 885.34 hours \\
Avg. Duration/Video & $\sim$52 minutes \\
Domains             & 17 \\
Total Word Tokens   & 5,832,680 \\
Unique Vocabulary   & 181,648 \\
Avg. WPM            & 114.81 \\
License             & CC-BY-4.0$^\ast$ \\
\bottomrule
\end{tabular}
\caption{Lipi-Ghor corpus summary. $^\ast$CC-BY-4.0 applies to our original contributions; caption-derived transcripts are for research use only (\S\ref{sec:ethics}).}
\label{tab:summary}
\end{table}

\subsection{Linguistic Analysis}

\paragraph{Speaking Rate.}
Average WPM varies widely across domains, from 170.17 in Service/Promo down to 59.87 in Regional/Dialect, spanning a broad range of speech styles relevant to ASR robustness.

\paragraph{Vocabulary and OOV.}
The corpus contains 181,648 unique word types. OOV rates vary sharply by domain: audiobooks reach 50.35\% (literary vocabulary), while news and political content fall to 15--16\% (constrained formal register).

\paragraph{Code-Switching.}
262 utterances (0.026\% of segments) contain Latin-script characters from natural Bengali--English code-switching. These are flagged in metadata but retained, as code-switching is a realistic feature of Bengali speech.

\paragraph{Transcription Quality.}
Transcriptions come from YouTube caption tracks, a mix of manually contributed and auto-generated captions, and are therefore weakly supervised. Lipi-Ghor is a \emph{pre-training corpus} rather than an evaluation benchmark.

\subsection{Conversational Structure}
\label{sec:diarstats}

Because every recording is diarized, we can characterize the corpus's conversational structure beyond raw duration.
Table~\ref{tab:diarstats} reports aggregate statistics over all 1,019 recordings.
Of the 885.34 hours of audio in the released corpus,
639 hours (72.2\%) are covered by at least one detected
speaker segment, while the remaining 246.34 hours
contain silence, music, ambient sound, or other
non-speech regions.

Speaker density is bimodal. The median recording has 4 speakers, but 25.9\% are single-speaker narration such as audiobooks and sermons, while 31\% carry ten or more speakers, mostly talkshows and debates.
Segments are also short, with a median of 1.54\,s and 62.6\% falling under two seconds, because diarization decomposes long recordings into turn-level chunks before ASR ever sees the audio.
This bears on the duration analysis in \S\ref{sec:zeroshot}, where segmentation has already removed most of the long-form difficulty.
Per-domain conversational statistics and the full speaker-density and segment-duration distributions appear in Appendix~\ref{app:stats}.

\begin{table}[t]
\centering
\small
\begin{tabular}{lr}
\toprule
\textbf{Property} & \textbf{Value} \\
\midrule
Diarized recordings      & 1,019 \\
Speaker-attributed segments & 984,647 \\
Speaker labels (all recordings) & 9,056 \\
\midrule
Speech / non-speech      & 639\,h / 243\,h \\
Speech ratio (mean)      & 0.76 \\
\midrule
Speakers per rec.\ (median, IQR) & 4 \ (1--12) \\
Single-speaker recordings & 25.9\% \\
Recordings with $\geq$10 speakers & 31.0\% \\
\midrule
Speaker turns per minute (mean) & 14.2 \\
Median segment duration  & 1.54\,s \\
Segments under 2\,s      & 62.6\% \\
\midrule
Detected overlap (median) & 0.31\% \\
Total overlapped speech  & 11.6\,h \\
\bottomrule
\end{tabular}
\caption{Conversational structure of the diarized corpus. Speech ratio is the fraction of each recording covered by at least one speaker turn, averaged over recordings.}
\label{tab:diarstats}
\end{table}
Table~\ref{tab:domstruct} gives the per-domain conversational statistics summarized in \S\ref{sec:diarstats}, obtained by joining the diarization output to the channel metadata on video ID (all 1,019 recordings match).
The domains fall into three groups.
Narration domains such as audiobook, education, and health are mostly single-speaker, turn slowly, and produce the longest segments.
Conversational domains such as talkshow, podcast, vlog, and comedy sit at 20--25 turns per minute with few solo recordings and high speech ratios.
Performance domains such as movie, natok, and cartoon show the highest speaker counts and the shortest segments, with speech ratios near 0.56 because music and sound effects occupy much of the audio.
Waz is a useful check on the diarizer: sermons are near-monologue, and the model agrees, giving a median of two speakers and a dominant-speaker share of 0.94.

\begin{table}[h]
\centering
\small
\setlength{\tabcolsep}{3.4pt}
\begin{tabular}{lrrrrr}
\toprule
\textbf{Domain} & \textbf{Spk} & \textbf{Solo} & \textbf{Turn} & \textbf{SR} & \textbf{Seg} \\
\midrule
Audio-book       &  1 & 62 &  3.4 & 0.74 & 1.83 \\
Talk-show        &  5 &  2 & 20.1 & 0.86 & 2.38 \\
Movie            & 24 & 14 & 20.0 & 0.58 & 0.95 \\
Natok            & 18 &  2 & 21.6 & 0.56 & 1.14 \\
Podcast          &  3 & 10 & 21.9 & 0.85 & 1.73 \\
Waz              &  2 & 32 &  5.8 & 0.84 & 2.21 \\
Music            &  3 & 14 & 17.4 & 0.71 & 2.12 \\
Comedy/Satire    &  7 &  5 & 22.0 & 0.73 & 1.54 \\
Vlog             &  9 &  0 & 25.1 & 0.70 & 1.22 \\
Education        &  1 & 64 &  3.8 & 0.90 & 3.63 \\
Political        &  3 & 17 & 11.5 & 0.90 & 2.56 \\
Cartoon          & 15 &  0 & 28.5 & 0.75 & 1.92 \\
Commentary       &  2 & 38 &  7.3 & 0.66 & 2.82 \\
News             &  5 & 14 & 18.1 & 0.84 & 2.82 \\
Regional/Dialect &  3 & 25 & 26.5 & 0.64 & 1.11 \\
Health           &  1 & 50 & 12.1 & 0.90 & 3.39 \\
Service/Promo    &  2 &  0 & 27.3 & 0.93 & 2.61 \\
\bottomrule
\end{tabular}
\caption{Conversational structure by domain. Spk = median speakers per recording; Solo = \% of single-speaker recordings; Turn = mean speaker turns per minute; SR = mean speech ratio; Seg = median segment duration (s). Ordered by total duration.}
\label{tab:domstruct}
\end{table}

\begin{figure}[t]
\centering
\includegraphics[width=\columnwidth]{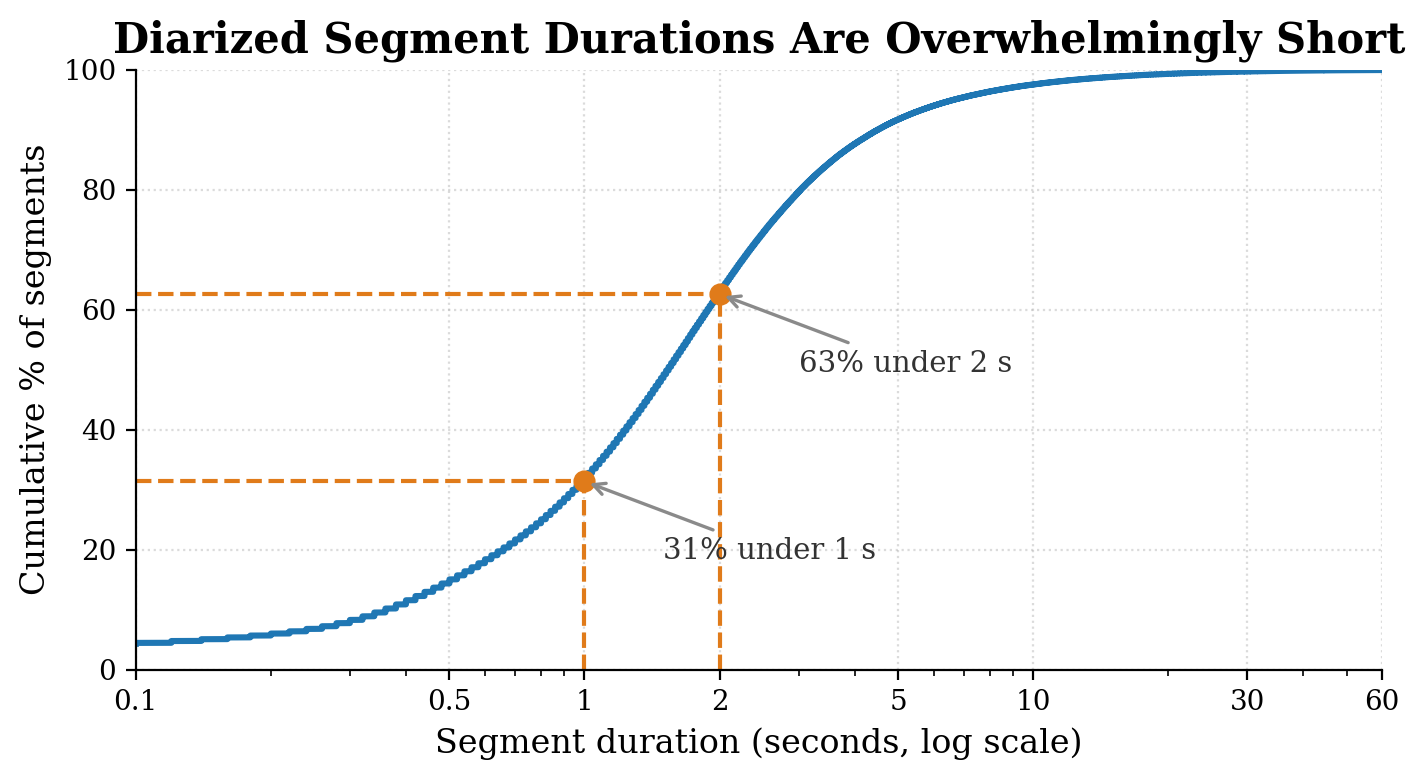}
\caption{Cumulative distribution of diarized segment durations (log-scaled x-axis). Nearly two-thirds of the 984,647 segments are shorter than two seconds.}
\label{fig:segcdf}
\end{figure}

Figure~\ref{fig:segcdf} shows the segment-duration distribution underlying the median of 1.54\,s reported in Table~\ref{tab:diarstats}.
The curve is steep throughout the sub-second range and flattens well before ten seconds, indicating that the diarization process converts long-form recordings into predominantly short speaker segments before ASR processing, even though the original source recordings average roughly 52 minutes in duration.

\section{Gold Evaluation Split}
\label{sec:gold}

We release a human-annotated gold split for evaluation.
The split comprises 500 clips drawn by stratified sampling across all 17 domains, with per-domain quotas so that even small domains (regional dialect, health) receive enough clips for meaningful per-domain estimates.
Seventeen volunteer native Bengali-speaking annotators produced approximately 30 hours of manual annotation: correcting caption-derived transcripts against the audio.
The gold split therefore carries both human transcriptions and human speaker labels, held out from training data, removing the main weakness of caption-referenced evaluation.

The gold split provides human-verified transcript references for ASR evaluation. The released diarization output remains automatically generated and is intended to support future diarization analysis (Section~\ref{sec:diarlimits}).

Table~\ref{tab:zeroshot} reports global WER and CER on the held-out test split, where all ASR systems are evaluated using the same reference transcripts from youtube captions.
These results provide a consistent comparison of ASR performance under the same evaluation setting. Differences across systems therefore reflect model behavior on the held-out test data rather than differences in reference quality.
The diarization analysis further highlights the limitations of automatic speaker segmentation discussed in \S\ref{sec:diarlimits}. In particular, over-segmentation and false speaker assignments can affect speaker-attributed transcript quality, especially in domains with frequent speaker changes.

Table~\ref{tab:gold_domain} compares the weakly supervised
YouTube transcripts with the human-corrected gold references
across different content domains.

\begin{table}[t]
\centering
\small
\setlength{\tabcolsep}{4pt}
\begin{tabular}{lrr}
\toprule
\textbf{Domain} & \textbf{WER} & \textbf{CER} \\
\midrule
Cartoon          & 28.57 & 25.85 \\
Vlog             & 14.17 &  4.07 \\
Commentary       & 10.81 &  5.33 \\
News             & 10.34 &  6.57 \\
Movie            &  7.32 &  5.28 \\
Waz              &  6.51 &  3.88 \\
Music            &  6.20 &  4.32 \\
Natok            &  5.88 &  3.43 \\
Podcast          &  3.81 &  2.81 \\
Talk-show        &  3.52 &  1.96 \\
Audio-book       &  2.57 &  1.12 \\
Comedy/Satire    &  0.00 &  0.00 \\
\bottomrule
\end{tabular}

\caption{
Domain-wise disagreement between the weakly supervised
YouTube transcripts and the human-corrected gold references.
WER and CER are corpus-level percentages calculated separately
for each domain.
}
\label{tab:gold_domain}
\end{table}

\section{Baseline Experiments}
\label{sec:zeroshot}

\begin{figure}[t]
\centering
\includegraphics[width=\columnwidth]{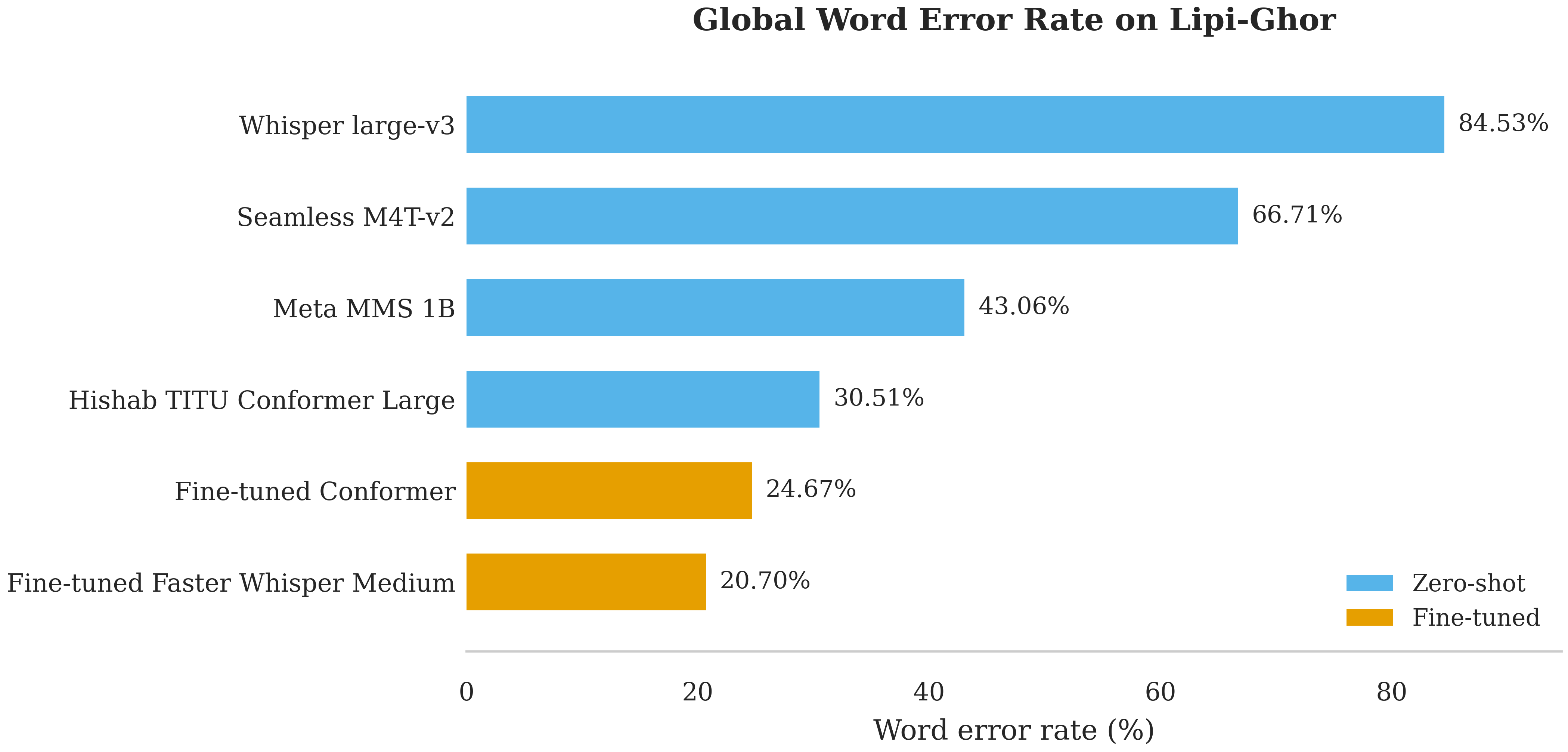}
\caption{Global WER for the four zero-shot systems (blue) and two fine-tuned models (orange). Fine-tuned Faster Whisper Medium reaches 20.70\%, below the best zero-shot system (Hishab TITU, 30.51\%).}
\label{fig:globalwer}
\end{figure}

Figure~\ref{fig:globalwer} places all six systems on one axis, the four zero-shot models alongside the two we fine-tuned on Lipi-Ghor.
The ordering is the main result: fine-tuning a mid-sized model on this corpus beats every off-the-shelf system we tested, including the Bengali-specific one.

\subsection{Zero-Shot Evaluation}

Table~\ref{tab:zeroshot} reports four ASR systems evaluated zero-shot on Lipi-Ghor using YouTube captions as reference.
All hypotheses and references are passed through an identical Bengali text-normalization pipeline (Unicode NFC, punctuation and digit standardization) before scoring.

\begin{table}[t]
\centering
\small
\begin{tabular}{lrr}
\toprule
\textbf{Model} & \textbf{WER} & \textbf{CER} \\
\midrule
Hishab TITU Conformer-L & 30.51 & 18.23 \\
Meta MMS 1B             & 43.06 & 21.16 \\
Seamless M4T-v2         & 66.71 & 45.54 \\
Whisper large-v3        & 84.53 & 72.92 \\
\bottomrule
\end{tabular}
\caption{Zero-shot ASR performance on Lipi-Ghor (\%).}
\label{tab:zeroshot}
\end{table}

The WER spread (30.51\%--84.53\%) shows both how hard the corpus is and how much Bengali-specific pre-training matters.
Whisper large-v3's very high error reflects its multilingual but English-centric training distribution and known weakness on Bengali script.
These figures overestimate true error against noisy captions; Section~5 quantifies this reference noise using the human-corrected gold split.

\begin{figure*}[t]
\centering
\includegraphics[width=\textwidth]{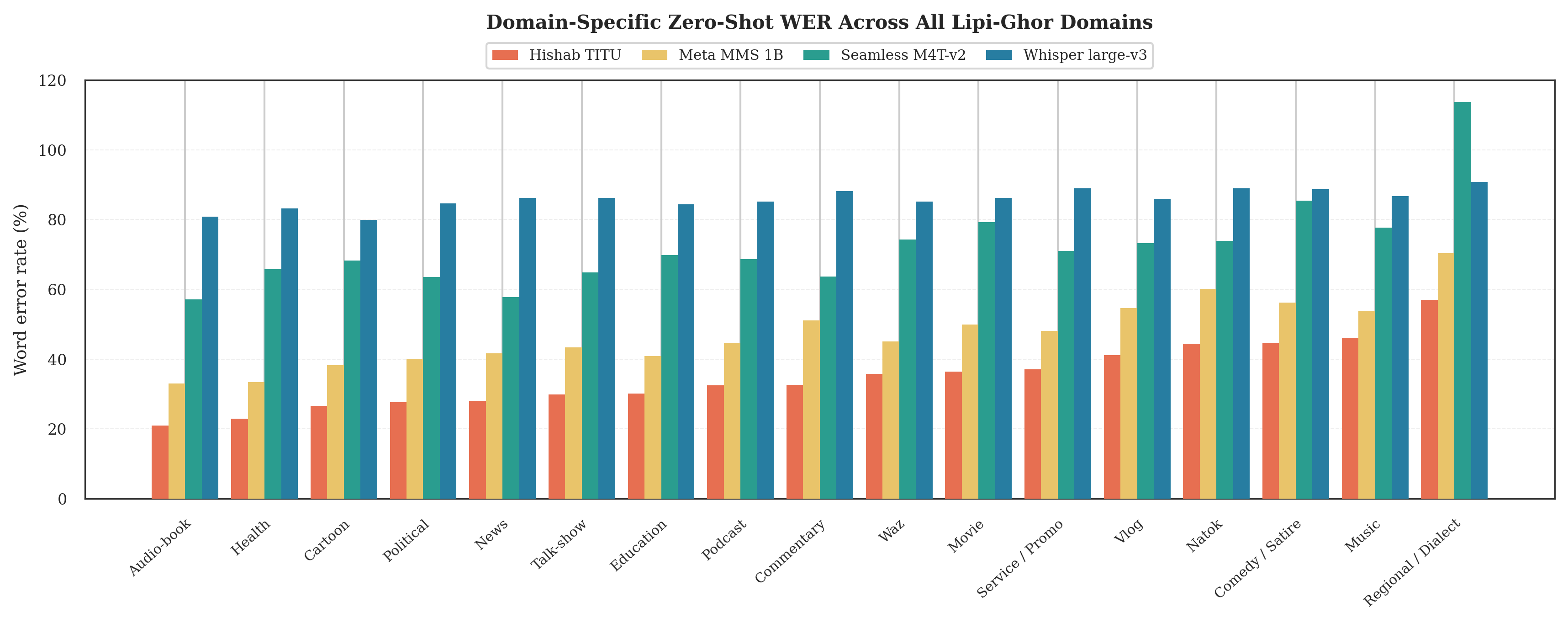}
\caption{Zero-shot WER across all 17 Lipi-Ghor domains, sorted by Hishab TITU's error rate. The three Bengali-capable systems degrade together as speech becomes more spontaneous and dialectal; Whisper large-v3 fails near-uniformly across domains. Table~\ref{tab:domainwer} lists a representative subset.}
\label{fig:domainwer}
\end{figure*}

\paragraph{Domain-Level Degradation.}
Table~\ref{tab:domainwer} shows domain-specific WER for representative domains; Figure~\ref{fig:domainwer} extends this to all seventeen.
All models follow a consistent degradation pattern: lowest WER on audiobooks and news, highest on emotive drama, comedy, and regional dialects.
Seamless M4T-v2 exceeds 100\% WER on Regional/Dialect, indicating near-complete failure.
Domains are sorted by TITU's error rate, and the other three curves rise with it almost monotonically, which suggests the difficulty ordering is a property of the audio rather than of any one model's training data.
Whisper large-v3 is the exception, staying near 85\% almost everywhere and failing at a level largely independent of domain.

\begin{table}[t]
\centering
\small
\begin{tabular}{lrrrr}
\toprule
\textbf{Domain} & \textbf{TITU} & \textbf{MMS} & \textbf{SM4T} & \textbf{Wh-v3} \\
\midrule
Audio-book       & 21.02 & 33.04 &  57.09 & 80.87 \\
News             & 28.03 & 41.68 &  57.83 & 86.23 \\
Talk-show        & 29.88 & 43.38 &  64.81 & 86.17 \\
Waz              & 35.75 & 45.02 &  74.27 & 85.15 \\
Natok            & 44.37 & 60.14 &  73.94 & 89.01 \\
Comedy/Satire    & 44.56 & 56.18 &  85.37 & 88.64 \\
Regional/Dialect & 57.01 & 70.39 & 113.70 & 90.74 \\
\bottomrule
\end{tabular}
\caption{Domain-specific WER (\%) for representative domains. SM4T = Seamless M4T-v2.}
\label{tab:domainwer}
\end{table}

\begin{figure}[t]
\centering
\includegraphics[width=\columnwidth]{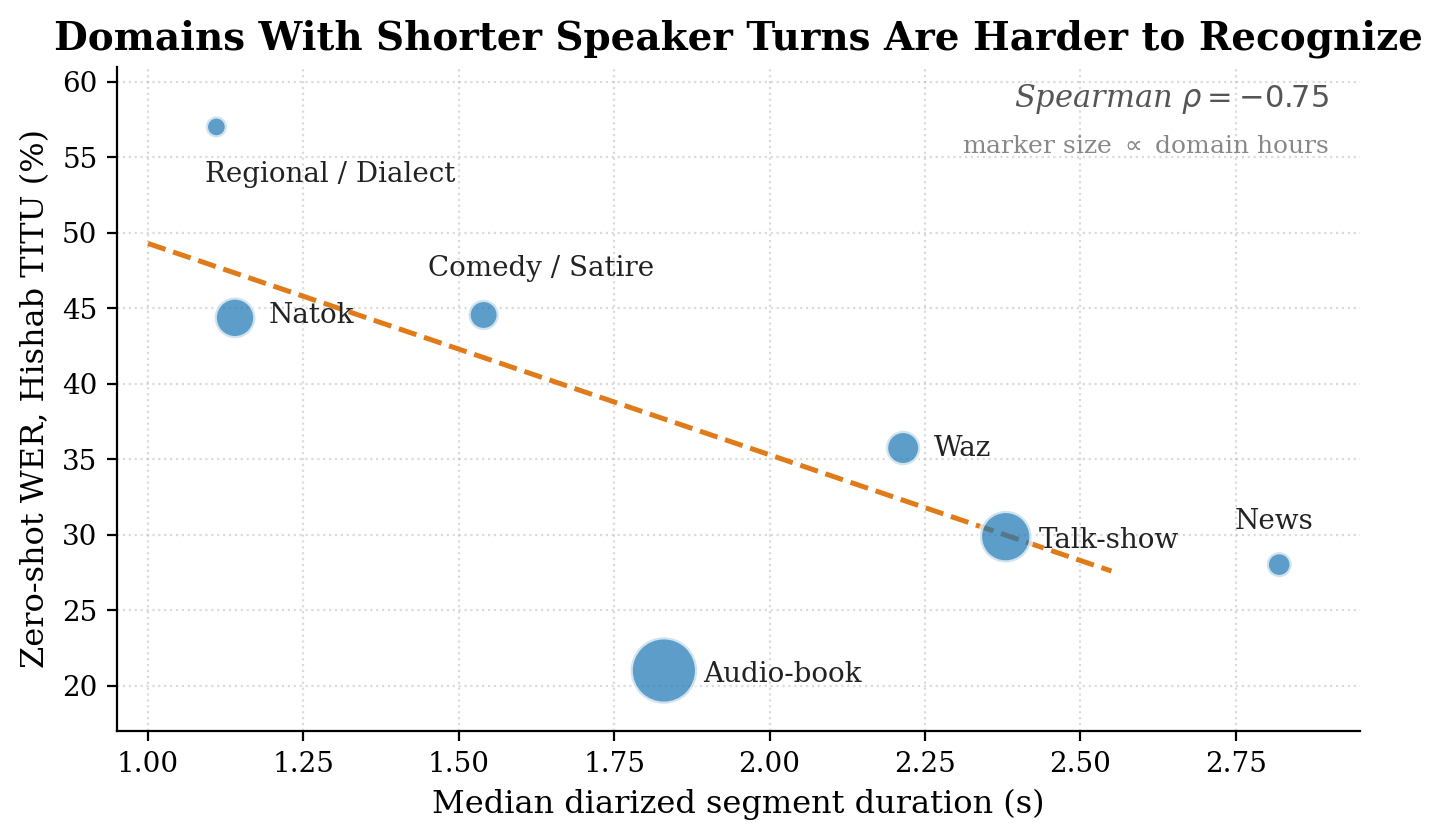}
\caption{Median diarized segment duration vs.\ zero-shot WER (Hishab TITU) across the seven domains of Table~\ref{tab:domainwer}. Marker area proportional to domain hours.}
\label{fig:turnwer}
\end{figure}

\paragraph{Structure and Difficulty.}
Across the seven domains in Table~\ref{tab:domainwer}, domains with the shortest diarized segments also have the highest error (Spearman $\rho = -0.75$; Figure~\ref{fig:turnwer}).
Audiobooks sit at one extreme with 1.83\,s median segments and 21.0\% WER, and Natok and Regional/Dialect at the other with roughly 1.1\,s segments and 44--57\% WER.
We want to be careful about how much weight this carries.
The relationship is suggestive rather than established, since it rests on seven points and weakens when the smallest domains are dropped, and turn rate on its own does not survive extension to all seventeen domains ($\rho = +0.42$, $p = 0.09$).
Cartoons are the clearest counterexample: they turn faster than any other domain at 28.5 turns per minute, yet remain easy to recognize because the speech is clearly enunciated.
What the correlation does support is a weaker and more useful claim, that fragmented turn-taking marks the domains where Bengali ASR struggles, and that segment duration is a cheap proxy for finding them without reference transcripts.

\paragraph{Utterance-Length Effects.}
Table~\ref{tab:duration} shows WER by utterance duration quartile.
The effect is modest, because diarization has already segmented long recordings into short turns and absorbed most of the long-form difficulty.
Only the very-long quartile shows a consistent uptick (TITU: 28.5\% $\to$ 32.6\%).

\begin{table}[t]
\centering
\small
\begin{tabular}{lrrrr}
\toprule
\textbf{Quartile} & \textbf{TITU} & \textbf{MMS} & \textbf{SM4T} & \textbf{Wh-v3} \\
\midrule
Short     & 30.54 & 41.98 & 66.11 & 85.01 \\
Medium    & 29.14 & 41.79 & 64.49 & 84.25 \\
Long      & 28.46 & 40.74 & 63.55 & 84.00 \\
Very Long & 32.60 & 45.53 & 70.13 & 84.93 \\
\bottomrule
\end{tabular}
\caption{WER (\%) by utterance duration quartile.}
\label{tab:duration}
\end{table}

\subsection{Fine-Tuned Baselines}

Table~\ref{tab:finetuned} presents a downstream utility evaluation of ASR models fine-tuned using Lipi-Ghor data. The evaluated models and their fine-tuning configurations follow the training setups described in \citet{hasan2026make}. Both models were trained using approximately 58 hours of Lipi-Ghor data under comparable data conditions.  The corresponding training subset is
publicly available at \href{https://huggingface.co/datasets/Sanjidh090/Train_lighten_check_bangla} {\texttt{Train\_lighten\_check\_bangla}}, together with the associated model checkpoints and training resources.
\textbf{Model A} (Fast Conformer) achieves 24.67\% WER and 15.56\% CER globally.
\textbf{Model B} (Faster Whisper Medium) achieves 20.70\% WER and 11.90\% CER globally.
On FLEURS Bengali, Model B scores 21.7\% and Model A 20.03\%, consistent with prior Bengali Whisper fine-tuning results.
For reference, Moonshine-Base fine-tuned with a transplanted BanglaBERT WordPiece vocabulary \citep{hasan2026moonshine} achieves 21.54\% WER and 10.79\% CER.
These gains over zero-shot baselines indicate that Lipi-Ghor is effective training data despite its weak supervision.
However, external benchmark results should be interpreted cautiously, as public models may have prior exposure to related Bengali resources.

\begin{table}[t]
\centering
\small
\begin{tabular}{lrr}
\toprule
\textbf{Model} & \textbf{WER} & \textbf{CER} \\
\midrule
Fast Conformer (A)          & 24.67 & 15.56 \\
Faster Whisper Med. (B)     & 20.70 & 11.90 \\
Moonshine-Base Bengali$^\dagger$ & 21.54 & 10.79 \\
\bottomrule
\end{tabular}
\caption{Fine-tuned model performance on Lipi-Ghor (\%). $^\dagger$\citet{hasan2026moonshine}; RTF 0.0053.}
\label{tab:finetuned}
\end{table}

\subsection{Gold-Split Evaluation}

Table~\ref{tab:gold_domain} provides an analysis of weak supervision quality on the 500-clip human-corrected split. It compares YouTube caption transcripts against human-corrected references across domains using WER and CER. These values measure transcript disagreement and should not be interpreted as ASR model error.

The level of transcript disagreement varies across domains. Some domains show higher differences between YouTube-generated transcripts and human corrections due to variations in caption quality and speech characteristics. Therefore, these values indicate reference transcript noise rather than ASR system performance.

\section{Discussion}

\paragraph{What the Diarization Gets Wrong.}
\label{sec:diarlimits}
Two patterns in the diarizer's output are artifacts rather than facts about Bengali speech.

First, the annotation never contains more than two simultaneous speakers: 682 recordings peak at two concurrent speakers, 337 at one.
Genuine talkshow and debate audio contains moments where three or more people talk over each other; this ceiling comes from Precision-2's output representation.
The overlap figures in Table~\ref{tab:diarstats} should be read as detected overlap under a two-speaker cap.

Second, detected overlap \emph{falls} as speaker count rises, which inverts what multi-party conversation would predict.
The likely cause is over-segmentation: when the diarizer splits a single voice across many labels, those fragments are consecutive rather than concurrent, so overlap drops.
Supporting evidence: 21.8\% of all 9,056 speaker labels account for under ten seconds of speech each, and the 105-label recording has a dominant-speaker share of 0.11.
Speaker counts are therefore reliable at the low end and inflated at the high end; the median of 4 is more trustworthy than the mean of 8.9.

\paragraph{Pipeline Generalizability.}
The construction methodology is not Bengali-specific by design.
Any language with a substantial YouTube presence and native-language caption tracks can serve as a corpus source.
We have only validated it on Bengali, so cross-lingual transfer is a hypothesis rather than a demonstrated result.
The bottleneck is domain-aware channel discovery: in our collection, 3--5 rich channels per domain collected 20--50 hours, suggesting comparable scales are achievable for other South Asian or low-resource languages with modest curation effort.

\section{Conclusion}

We presented Lipi-Ghor, an 885-hour weakly supervised Bengali speech
corpus spanning 17 domains, together with a scalable YouTube-based
construction pipeline. Experiments reveal consistent degradation on
emotive, dialectal, and long-form speech, while fine-tuning demonstrates
the corpus's utility for Bengali ASR. We release the corpus metadata,
annotations, pipeline code, and human-annotated gold evaluation split.
\section*{Limitations}

The most significant limitation is weak supervision: transcripts are derived from YouTube caption tracks, have not been fully human-verified, and systematically differ from human transcriptions in punctuation, digit normalization, and code-switched speech handling.
Speaker labels over the corpus are produced by Precision-2 and are not human-verified; therefore, the “Speaker” dimension of SSTT should be regarded as automatically generated corpus metadata rather than a manually validated diarization benchmark.
Precision-2 is a commercial, API-gated model, so the diarization stage is not fully open; we release its output so downstream use does not require re-running it.
Regional dialect coverage is sparse (4 videos, 1.29 hours), insufficient for dialect-robust training.
Exact reproducibility is constrained by YouTube's evolving API rate policies and the partially manual collection process; we mitigate this by releasing video identifiers and pipeline scripts.
\section{Ethical Considerations}
\label{sec:ethics}

\paragraph{Data Source, Copyright, and Licensing.}
We do not redistribute raw audio. We release video identifiers and pipeline scripts so users reconstruct audio locally, following the video-ID release model of \citet{bengaliloop2025} and \citet{galvez2021peoples}.
Our original contributions (video-ID lists, diarization output, pipeline code, and gold annotations) are released under CC-BY-4.0.
Caption-derived transcripts remain subject to the rights of the original uploaders and are distributed for non-commercial research use only.

\paragraph{Data Removal Policy.}
Rights holders, uploaders, or individuals whose voices appear in source videos may request removal by supplying video identifiers.
On a valid request we remove the corresponding entries, publish an updated dataset version, and record removed identifiers in a public changelog.

\paragraph{Speaker Consent and Privacy.}
Speech comes from public YouTube channels whose creators published the content themselves.
We restrict the release to research use and prohibit speaker re-identification and surveillance applications.
Speaker labels are local to each video and do not support cross-video re-identification.

\paragraph{Intended Use.}
Lipi-Ghor is intended for ASR pre-training, spoken language modeling, and multilingual NLP research. It is not intended for speaker re-identification, surveillance, or high-stakes production ASR without additional human verification.

\section*{Acknowledgments}

The authors are grateful to Almighty Allah for His guidance and blessings throughout this work.
\bibliography{custom}

\clearpage

\appendix

\section{Extended Dataset Statistics}
\label{app:stats}

\begin{table}[H]
\centering
\small
\begin{tabular}{lrr}
\toprule
\textbf{Domain} & \textbf{Videos$^\ast$} & \textbf{Hours} \\
\midrule
Audio-book       & 277 & 254.37 \\
Talk-show        & 176 & 142.40 \\
Movie            &  74 & 122.18 \\
Natok (Drama)    &  95 &  78.66 \\
Podcast          &  62 &  57.59 \\
Waz (Sermons)    &  66 &  47.89 \\
Music            &  37 &  31.91 \\
Comedy/Satire    &  43 &  31.35 \\
Vlog             &  42 &  27.13 \\
Education        &  39 &  25.14 \\
Political        &  29 &  19.35 \\
Cartoon          &  23 &  16.88 \\
Commentary       &  13 &  14.51 \\
News             &  35 &  12.88 \\
Regional/Dialect &   4 &   1.29 \\
Health           &   2 &   1.14 \\
Service/Promo    &   2 &   0.67 \\
\midrule
\textbf{Total}   & \textbf{1{,}019} & \textbf{885.34} \\
\bottomrule
\end{tabular}
\caption{Domain distribution of Lipi-Ghor. $^\ast$Per-domain counts are raw video counts before cross-domain deduplication (1,019 total); deduplication yields 998 unique videos. Hours are on the deduplicated corpus.}
\label{tab:domains}
\end{table}
\begin{table}[h]
\centering
\caption{Distribution of YouTube Caption Types in Lipi-Ghor}
\label{tab:caption_types}
\begin{tabular}{lrr}
\hline
\textbf{Caption Type} & \textbf{Videos} & \textbf{Percentage} \\
\hline
Manually created captions & 102 & 10.22\% \\
Auto-generated captions   & 896 & 89.78\% \\
\hline
Total                     & 998 & 100.00\% \\
\hline
\end{tabular}
\end{table}

\section{The Annotation Toolkit}
\label{app:tool}

\begin{figure*}[t]
\centering
\includegraphics[width=\textwidth]{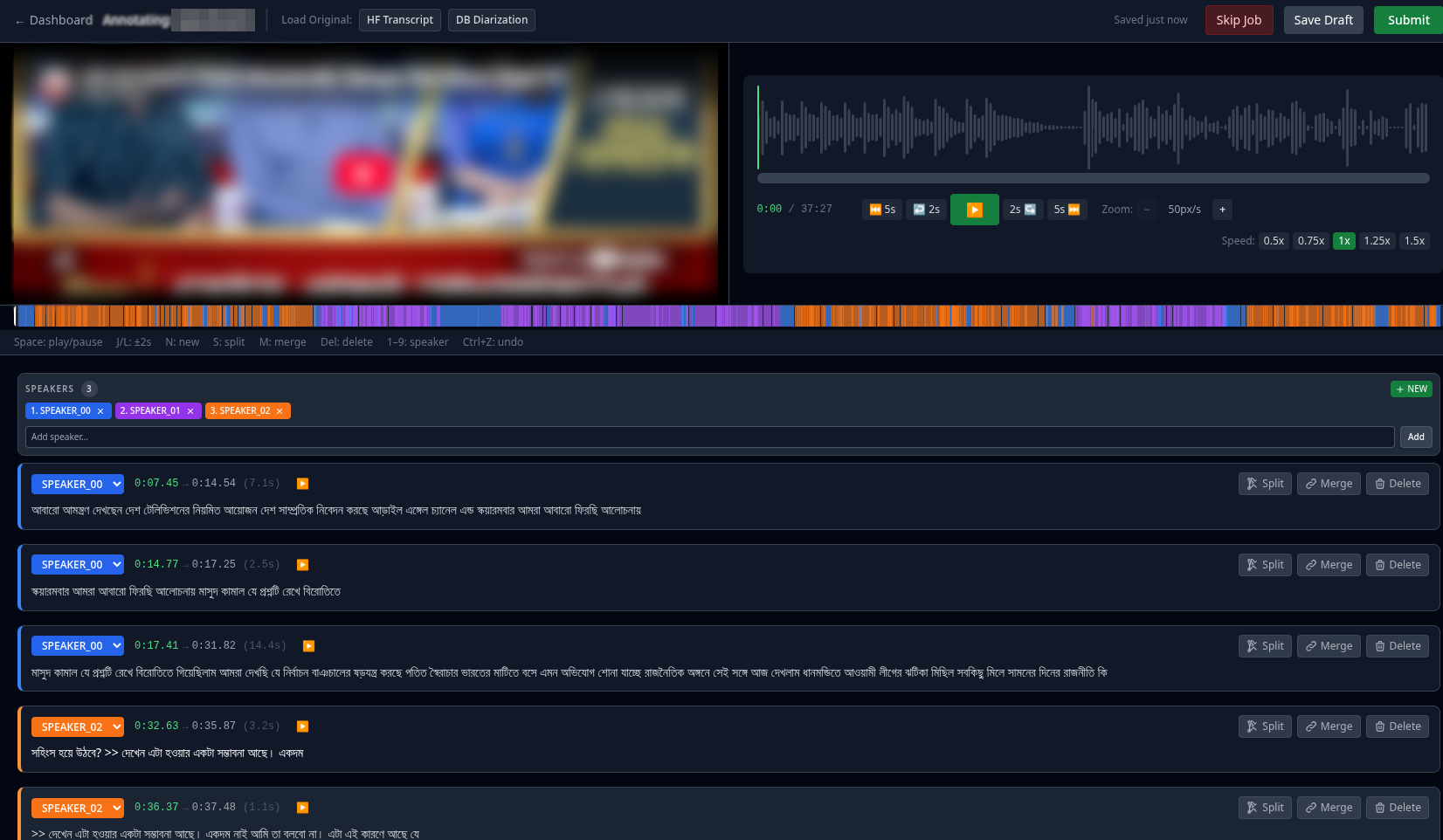}
\caption{The Lipi-Ghor annotation interface. Clockwise from top left: the embedded video player, the waveform with a per-speaker diarization ribbon beneath it, the speaker roster for the recording, and the segment list where each row carries a speaker assignment, timestamps, and an editable Bengali transcript. Annotators can split, merge, or delete segments and reassign speakers from the dropdown. Identifying details of the source video, including the video identifier and all channel branding, are masked for double-blind review; annotators saw the unmasked frame.}
\label{fig:anntool}
\end{figure*}

The 500-clip gold split (\S\ref{sec:gold}) was produced with a purpose-built web annotation tool rather than a general-purpose labeling platform.
We describe it here because the design determines what the gold labels can and cannot be trusted to contain, and because the corpus release includes the tool's export scripts.
Figure~\ref{fig:anntool} shows the annotation screen as a volunteer sees it.

\vspace{20pt}
\paragraph{Interface.}
Each annotation job loads one recording together with its automatic annotations, so annotators correct existing output rather than transcribing from scratch.
The two ``Load Original'' controls re-fetch the caption-derived transcript and the Precision-2 diarization, which lets an annotator revert to the machine baseline if an edit goes wrong.
The waveform supports variable-speed playback from 0.5$\times$ to 1.5$\times$ and 2\,s or 5\,s jumps, and the coloured ribbon under it shows the current speaker assignment across the whole recording, so an annotator can see turn structure at a glance instead of inferring it row by row.
The segment list below is the editing surface: every row exposes the speaker label, the start and end timestamps, and the Bengali transcript, and rows can be split at the playhead, merged with a neighbour, or deleted outright.
Keyboard shortcuts cover the frequent operations, including play/pause, $\pm$2\,s seeking, split, merge, delete, undo, and direct speaker assignment via the number keys.
Work is saved as a draft continuously, and a job is submitted only once the annotator has reviewed every segment.
\vspace{20pt}
\paragraph{Why the Video Is Retained.}
The tool deliberately keeps the source video visible, which is a departure from audio-only transcription interfaces.
The reason is that our annotators verify speaker turns as well as transcripts, and speaker attribution from audio alone is unreliable in exactly the domains where the corpus is densest.
Talkshow panels often place several speakers of the same gender in similar acoustic conditions, and natok scenes may involve several actors with deliberately similar delivery; in both cases voice quality is a weak cue.
Seeing who is on screen, whose lips are moving, and how the camera frames a turn resolves attributions that the waveform leaves genuinely ambiguous.
Removing the player would have made the gold speaker labels less reliable, which would defeat their purpose as a reference for measuring diarization error.
The trade-off is that visual cues can also mislead, since an off-screen speaker may be talking while the camera holds on someone else, so annotators were instructed to treat the audio as authoritative and the video as a disambiguation aid rather than the reverse.

\paragraph{Provenance and Export.}
Each submitted job records the annotator, the timestamps of the edits,
and the machine baseline the annotator started from, so a corrected
segment can always be compared against the automatic output it replaced.
This provenance enables future fine-grained analysis of diarization
errors at the segment level and supports reproducible evaluation against
human-corrected annotations.
Completed jobs are exported to the same SSTT schema used by the bulk
corpus, so the gold split and the weakly supervised corpus share one
format and the gold clips can be substituted for their noisy counterparts
without any change to a training pipeline.

\section{Diarization Output Licensing}
\label{app:licensing}

Speaker-diarization annotations were generated using the commercial
pyannoteAI Precision-2 service. Under pyannoteAI's Terms of Use, intellectual
property rights in results produced by the service are assigned to the
customer and may be reproduced, modified, reused, and transferred to third
parties. Accordingly, we release the generated Lipi-Ghor diarization outputs
under CC-BY-4.0. This license applies only to the generated annotations and
does not apply to the proprietary Precision-2 model or the pyannoteAI service
itself.

\end{document}